\documentclass[conference]{IEEEtran}
\IEEEoverridecommandlockouts
\usepackage{cite}
\usepackage{amsmath,amssymb,amsfonts,bm}
\usepackage{algorithmic}
\usepackage{graphicx}
\usepackage{graphics}
\usepackage{textcomp}
\usepackage{xcolor}
\usepackage{booktabs}
\usepackage{svg}
\usepackage[shortcuts, acronym]{glossaries}
\usepackage{multirow}
\usepackage[normalem]{ulem}
\usepackage{diagbox}
\usepackage{hyperref}
\hypersetup{
    colorlinks,
    citecolor=black,
    filecolor=black,
    linkcolor=black,
    urlcolor=blue
}

\useunder{\uline}{\ul}{}

\def\BibTeX{{\rm B\kern-.05em{\sc i\kern-.025em b}\kern-.08em
    T\kern-.1667em\lower.7ex\hbox{E}\kern-.125emX}}

\makeglossaries % generates the acronym list
\newacronym{arima}{ARIMA}{Autoregressive Integrated Moving Average}
\newacronym{var}{VAR}{Vector Autoregressive Model}
\newacronym{gru}{GRU}{Gated Recurrent Unit}
\newacronym{cnn}{CNN}{Convolutional Neural Network}
\newacronym{rnn}{RNN}{Recurrent Neural Network}
\newacronym{mlp}{MLP}{Multi-layer Perceptron}
\newacronym{nlp}{NLP}{Natural Language Processing}
\newacronym{gnn}{GNN}{Graph Neural Network}
\newacronym{gcn}{GCN}{Graph Convolutional Network}
\newacronym{gat}{GAT}{Graph Attention Network}
\newacronym{stgcn}{STGCN}{Spatio-Temporal Graph Convolutional Networks}
\newacronym{gman}{GMAN}{Graph Multi-Attention Network}
\newacronym{dcrnn}{DCRNN}{Diffusion Convolutional Recurrent Neural Network}
\newacronym{mtgnn}{MTGNN}{Multivariate Time Series Graph Neural Network}
\newacronym{gts}{GTS}{Graph for Time Series}
\newacronym{lgf}{LGF}{Layer-wise Gated Fusion}
\newacronym{cnf}{CNF}{Conditional Neural Field}
\newacronym{rff}{RFF}{random Fourier features}
\newacronym{mae}{MAE}{Mean Absolute Error}
\newacronym{mape}{MAPE}{Mean Absolute Percentage Error}
\newacronym{smape}{sMAPE}{symmetric Mean Absolute Percentage Error}
\newacronym{rmse}{RMSE}{Root Mean Square Error}
\newacronym{seacnn}{SEACNN}{SEasonality-Aware Convolutional Neural Network}
\newacronym{seagnn}{SEAGNN}{SEasonality-Aware Graph Neural Network}

\DeclareMathOperator*{\argmin}{arg\,min}

\newcommand{\etal}{\textit{et al.}}

\def\Tableref#1{Table~\ref{#1}}
\def\Figref#1{Fig.~\ref{#1}}

\def\eqref#1{equation~\ref{#1}}
\def\Eqref#1{Equation~\ref{#1}}
\def\1{\bm{1}}

\def\vb{{\bm{b}}}

\def\vx{{\bm{x}}}

\def\vz{{\bm{z}}}

\def\mB{{\bm{B}}}

\def\mH{{\bm{H}}}

\def\mW{{\bm{W}}}
\def\mX{{\bm{X}}}
\def\mY{{\bm{Y}}}

\def\gD{{\mathcal{D}}}
\def\gE{{\mathcal{E}}}

\def\gG{{\mathcal{G}}}

\def\gL{{\mathcal{L}}}

\def\gN{{\mathcal{N}}}

\def\gV{{\mathcal{V}}}

\def\sE{{\mathbb{E}}}

\def\sR{{\mathbb{R}}}

\begin{document}

% This paper is prepared to submit to IEEE Trans. on Neural Networks and Learning Systems Special Issue on Graph Learning https://cis.ieee.org/images/files/Documents/call-for-papers/tnnls/CFP-TNNLS-Special_Issue_on_Graph_Learning.pdf

\title{
%Time Series Forecasting: Conditional Neural Fields Make Spatial-Temporal Neural Networks More Effective \\
A Generic Approach to Integrating Time into Spatial-Temporal Forecasting via Conditional Neural Fields \\
%Improving Time Series Forecasting by Integrating Time into Spatial-Temporal Forecasting 
}

\author{
    Minh-Thanh Bui, Duc-Thinh Ngo, Demin Lu, and Zonghua Zhang
    \thanks{M. T. Bui, D. Lu, and Z. Zhang are with Paris Research Center, Huawei Technologies France.}
    \thanks{D. T. Ngo is currently with IMT Atlantique, Nantes Université, École Centrale Nantes, CNRS, INRIA, LS2N, UMR 6004, F-44000 Nantes, France, and also with Orange Labs, France. He contributed to this project while working as a research intern at Huawei Paris Research Center.}

} % end author

\maketitle
\thispagestyle{plain}
\pagestyle{plain}
\begin{abstract}
Self-awareness is the key capability of autonomous systems, e.g., autonomous driving network, which relies on highly efficient time series forecasting algorithm to enable the system to reason about the future state of the environment, as well as its effect on the system behavior as time progresses. Recently, a large number of forecasting algorithms using either convolutional neural networks or graph neural networks have been developed to exploit the complex temporal and spatial dependencies present in the time series. While these solutions have shown significant advantages over statistical approaches, one open question is to effectively incorporate the global information which represents the seasonality patterns via the time component of time series into the forecasting models to improve their accuracy. This paper presents a general approach to integrating the time component into forecasting models. The main idea is to employ conditional neural fields to represent the auxiliary features extracted from the time component to obtain the global information, which will be effectively combined with the local information extracted from autoregressive neural networks through a layer-wise gated fusion module. Extensive experiments on road traffic and cellular network traffic datasets prove the effectiveness of the proposed approach.
\end{abstract}


\section{Introduction}
\label{sect:introduction}

% Time series forecasting has gained various applications, such as weather forecasting, 

Time series forecasting is a well-studied topic that has attracted tremendous efforts for various applications, e.g., weather temperature, electricity usage, traffic speed, mobile traffic usage.
%thus enabling an increasing level of connectivity and %intelligence in our world.  
%The data collected from these sensors are typically in the form %of time series, either univariate or multivariate, and are %analyzed for various purposes, including condition monitoring, %event detection, and forecasting, in a wide range of %applications, such as autonomous vehicles, energy optimization, %network management, and resource allocation.
In recent years, time series forecasting has become a core capability in autonomous systems (e.g., autonomous driving cars, autonomous driving networks), as it may contribute to improving the self-awareness of such systems and enable proactive and autonomous decisions. 
Specifically, time series forecasting methods can reason about future states of the environment and how they affect the system behavior over time \cite{bauer2020time}. 
Some conventional statistical approaches such as the vector auto-regressive model (VAR) and Gaussian process model (GP) have been widely used to explore the linear temporal dependency among variables for prediction. 
Deep learning models have recently drawn a lot of attention thanks to their capability of capturing non-linear patterns present in data and leveraging the advantages of big amounts of data.
While temporal dependency between variables serves as essential information for forecasting, other information like spatial properties should not be ignored, especially for those multivariate time series. 
For example, in road traffic prediction \cite{stgcn_2018, gman_2020}, both the historical records of sensors deployed in the road network and the road structure should be taken into account. 
%in road traffic forecasting consists in predicting the future %traffic speeds of each sensor in a road network, given their %historical traffic speeds with or without the underlying road %structure networks. 
%This is a challenging task due to the complex spatial and temporal dependencies among the time series and the non-stationary characteristics caused by traffic congestion or unexpected driving behaviors. 
To do that, a variety of spatio-temporal \ac{gnn} frameworks \cite{stgcn_2018, dcrnn_2018, gts_2021, gman_2020, mtgnn_2020} has been developed for simultaneously exploring spatial and temporal dependencies by taking advantages of deep learning networks and \ac{gnn}s.
%to jointly explore the spatial and temporal dependencies simultaneously during the model construction process have achieved great success in traffic prediction recently. 
\begin{figure}
    \centering
    \includegraphics[width=\columnwidth]{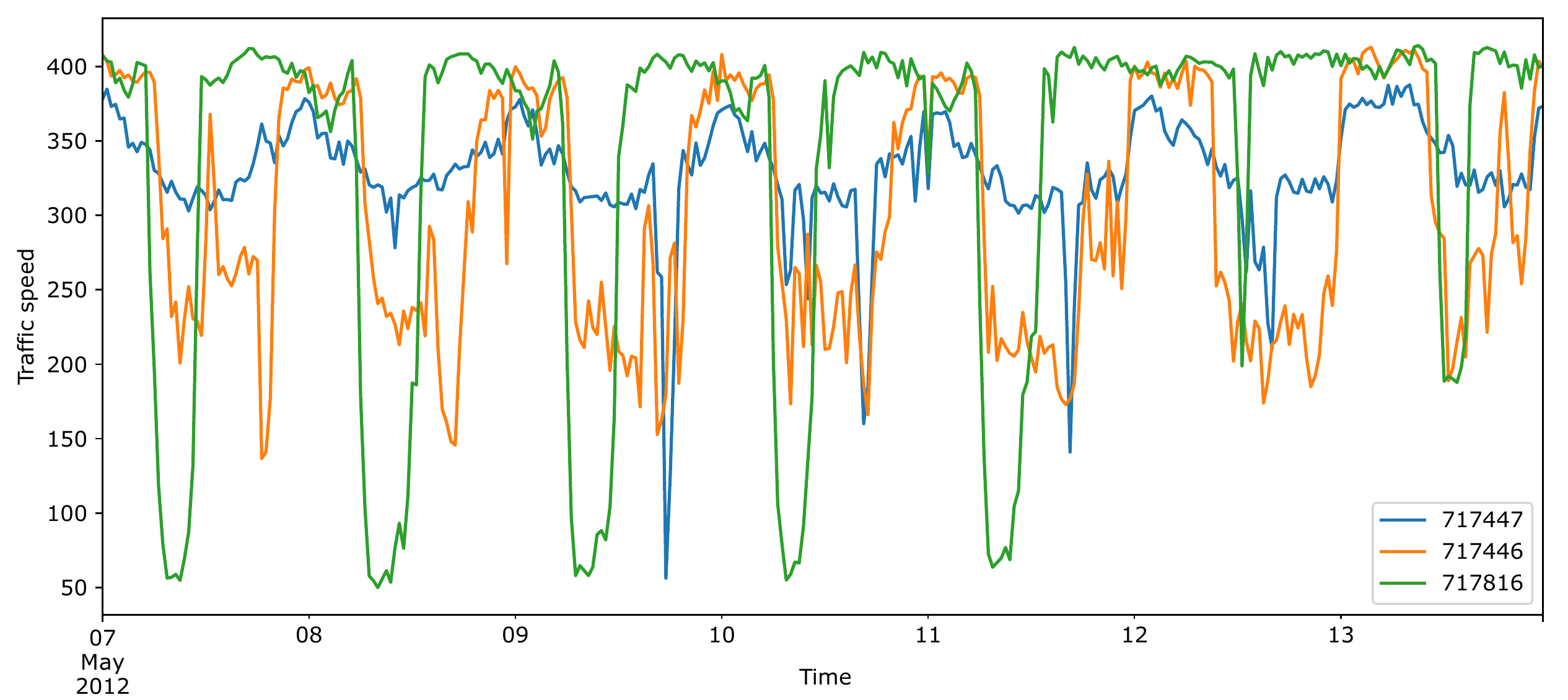}
    \caption{An example of raw traffic data in the METR-LA dataset. 
    The data are resampled to a 30-minute resolution for the sake of clarity. The traffic patterns exhibit variations between weekdays and weekends, with differing traffic speeds during midday and midnight compared to other periods throughout the day.}
    \label{fig:traffic_data}
\end{figure}

In addition, as shown in \Figref{fig:traffic_data}, seasonality over weeks and days is clearly present in road traffic data, which can be extracted and represented as a long-range time-dependent global feature.   
%This highlights the critical role of time as a component in traffic data analysis beyond just the numerical values themselves. 
%Seasonality can serve as a global feature that can be extracted from each time series and consists of solely time-dependent characteristics. 
%For traffic data, the global information can be the representative patterns occurring repeatedly over time. 
%Overall, the global information is important prior to the forecasting or regression task in question. 
As a matter of fact, the auto-regressive models usually use the separated subsequences of historical data to perform forecasting \cite{dcrnn_2018, mtgnn_2020, stgcn_2018, gman_2020}, thereby exhibiting a bias towards the recent historical data, although the non-stationary dynamics can be captured.
%the state-of-the-art solutions for the traffic prediction problems are based only on 
%Although such models are able to capture the non-stationary dynamics of the time series, they exhibit a bias towards the recent past data. 
In other words, the auto-regressive models mainly consider local features (i.e., numerical values) rather than global features (i.e., long-range time-dependent seasonality). 
%Hence, we postulate that by simultaneously utilizing both local and global information, it is possible to enhance the accuracy of traffic prediction. 
%While the local features derived from autoregressive models have been investigated extensively in the traffic prediction literature, the global features have received limited attention. 

Given these key observations, we are motivated to develop a novel method to improve time series forecasting performance by taking into account spatial-temporal properties, while preserving both local and global features.  
%Therefore, we propose to adopt a \ac{cnf} for the representation of the time component of the time series as the global features.  
To that end, we propose to use \ac{cnf} to represent the time component of time series as the global features. 
Specifically, the neural field is essentially a coordinate-based \acrfull{mlp} network that takes the auxiliary features of timestamps as input coordinates and outputs the temporal values. 
%Specifically, \ac{cnf} is used to represent the time component of time series as the global features. 
%The neural field is basically a coordinate-based \acrfull{mlp} network which takes the auxiliary features of timestamps as input and outputs the temporal values. 
In other words, if we consider a time series as a function of timestamps, a coordinate-based neural network can be used to approximate this function, enabling it to represent the global feature of a time series. 
In particular, \ac{cnf} may bring two major advantages. 
First, the neural field is invariant to the size of the time series, as it only takes timestamps as input, regardless of the length of the time series. 
Second, the neural field is lightweight by design, but it can represent various types of data, including images, 3D models, and audio \cite{vsitzmann, fourfeat, neural_fields}.

However, the application of \ac{cnf} is non-trivial, and we have to systematically study the following questions:
(1) which neural field architectures can best represent a time series? 
(2) how to encode {\em multiple} time series efficiently using neural fields? and 
(3) how to effectively combine the global features extracted from neural fields with the local features?  
As a result, the following contributions are delivered:  
\begin{itemize}
    \item We propose an effective approach consisting of \ac{cnf} and \ac{lgf} modules to integrate {\em time component} into forecasting models. 
    In particular, the \ac{cnf} module is used to represent auxiliary time features from multiple time series, while the \ac{lgf} module is used to combine the global features (i.e., daily and weekly seasonality) generated by the \ac{cnf} module with the local features obtained by the autoregressive models.
    \item We develop two novel forecasting models to integrate time component using our proposed approach into the inception forecasting algorithm and graph-based \ac{mtgnn} \cite{mtgnn_2020}, which are called \ac{seacnn} and \ac{seagnn} respectively. 
    \item We perform extensive experiments on two public road traffic datasets and two cellular mobile network traffic datasets to demonstrate that the two novel forecasting algorithms outperform their respective counterparts in terms of well-defined performance metrics, indicating the generalization capability of the proposed approach.
\end{itemize}

%This paper describes the development of a general framework for improving the accuracy of spatio-temporal forecasting algorithms. 

%As traffic of road networks \cite{dcrnn_2018, mtgnn_2020} and that of cellular mobile networks \cite{zhang2018citywide, sthgcn_2020, green5g} share similar characteristics, we use both of them as evaluation scenarios.  
%Traffic forecasting on road networks \cite{dcrnn_2018, mtgnn_2020} and cellular mobile networks \cite{zhang2018citywide, sthgcn_2020, green5g} are used as use cases to facilitate the development of the %algorithm. 
%In particular, to evaluate the generalization capability of our proposed framework, we first examine its performance with road traffic, and then extend the performance evaluation with cellular mobile networks. 

%As these two kinds of networks share similar characteristics, the paper first discusses traffic forecasting on road networks and then extends the work to cellular mobile networks to demonstrate the %generalization capability of the developed framework. 

The rest of this paper is organized as follows. 
Section \ref{sect:sota} surveys the related work, with a particular focus on the recent spatio-temporal forecasting algorithms. 
Our proposed approach and two novel forecasting algorithms (\ac{seacnn} and \ac{seagnn}) are described in detail in Section \ref{sect:methods}. 
Section \ref{sect:experiments} reports the experimental evaluation, including datasets, evaluation metrics, and comparative studies.
Finally, conclusion is drawn in Section \ref{sect:conclusions}.

\section{Related Work}
\label{sect:sota}

%\textbf{Problem Formulation:} 
Time series forecasting has been studied for decades, and a large set of algorithms has been developed so far. 
It is well recognized that traditional statistical methods (e.g., \ac{arima} and \ac{var}) are not effective in multivariate time series forecasting, as they can hardly capture non-linear inter- and intra-dependencies of data. 
Deep learning methods have been proved to do that better by leveraging the advantages of big data.  
The general methodology of deep learning methods is to design computational blocks to explore spatial and temporal properties of time series data and integrate these blocks into an end-to-end learning framework. 
In road traffic prediction, for example, a pre-defined graph can be constructed by using the structure of traffic networks based on sensor distance and traffic flow. 
Then a \ac{gnn} architecture can be developed to process information from neighboring nodes for capturing their spatial dependencies, which are then combined with the high-level temporal features from each node to eventually establish an end-to-end learning framework \cite{stgcn_2018, sthgcn_2020, dcrnn_2018, gman_2020, informer_2021, traversenet_2022}.  
%However, previous work mostly focuses on the spatial blocks and how to incorporate them into an end-to-end framework, while the design of the temporal blocks has not shown any significant progress. 
%We will review the most recent spatial-temporal \ac{gnn} forecasting methods that require a pre-defined graph structure or learn to construct the graph structure themselves.
%By taking into account the structure of traffic networks such as sensor distance, traffic flow, and prior knowledge, a pre-defined graph can be constructed. 
%Based on the predefined graph, the effort has been put into designing an effective \ac{gnn} architecture, i.e., spatial block, to process the information from neighboring nodes to %capture the spatial dependencies and combine them with the 
%high-level temporal features from each node extracted by temporal blocks in order to finally make an end-to-end learning framework \cite{stgcn_2018, sthgcn_2020, dcrnn_2018, %gman_2020, informer_2021, traversenet_2022}. 

In particular, Yu \etal \cite{stgcn_2018} proposed \ac{stgcn} framework by combining graph convolution and gated temporal convolution through spatio-temporal convolutional blocks, each of which consists of two temporal blocks and one spatial block in between. 
The temporal block is constructed using 1-D causal convolutional operators followed by gated linear units as a non-linearity. 
The spatial block takes a pre-defined graph and temporal features extracted for each node, and then applies graph convolutions to obtain spatial-state propagation from graph convolution through temporal convolutions. 
In \cite{dcrnn_2018}, the authors proposed \ac{dcrnn} to integrate temporal and spatial dependencies. 
Specifically, an \ac{rnn} framework built from \ac{gru}s was used to extract temporal features. 
In each \ac{gru} cell, a \ac{gnn} block was then used to extract spatial features. 
Instead of using ChebNet in \ac{stgcn} that only handles undirected graphs, a \textit{diffusion convolution} was applied to handle both directed and undirected graphs.

In addition, the authors of \cite{gman_2020} developed GMAN to perform traffic forecasting by employing a transformer framework that includes self-attention and cross-attention for enabling long-range forecasting. 
GMAN particularly takes into account both temporal attention and attention along the spatial axis. 
%While the \textit{temporal attention} block was inherited directly from the attention mechanism on sequential data \cite{vaswani2017}, the \textit{spatial attention} block was inspired from \ac{gat} \cite{gat}.
Similarly, Zhou \etal \cite{informer_2021} presented the Informer model for long-sequence time series forecasting. 
The ProbSparse self-attention mechanism and the distilling operation were developed to reduce computational complexity and memory usage in the vanilla Transformer \cite{vaswani2017}. 
% The ProbSparse self-attention mechanism aims to reduce the computational complexity by considering only the most important features of the input sequence, whereas the distilling operation aims to alleviate the memory usage by compressing the information of the self-attention module into a more compact representation.  
The authors also introduced a generative decoder, which, in combination with the self-attention distilling operation, to improve the efficiency of long-sequence predictions.
In \cite{traversenet_2022}, the authors proposed a spatial-temporal \ac{gnn} called TraverseNet for traffic forecasting that unified space and time as a single entity, which an objective to addressing the limitations of existing spatial-temporal neural networks by better capturing the complex relationships between space and time in traffic forecasting.  
Interestingly, the model captures the dependencies among nodes in a spatial-temporal graph by using a message traverse mechanism to exploit evolving spatial-temporal dependencies for each node. 

All the aforementioned work assumes that the prior traffic network structure and prior knowledge are readily available, which is however not always true. 
%, or the pre-defined graph is not precise nor helpful in improving the accuracy of the traffic prediction. 
It thus becomes necessary to automatically learn the graph structure \cite{bai2020adaptive, gts_2021, mtgnn_2020}. 
For example, in \cite{gts_2021}, Shang \etal formulated graph structure learning as a learning process of the graph distribution within a probabilistic graphical model, which is parameterized by neural networks and optimized jointly with the forecasting model. 
Specifically, the authors proposed to learn the graph structure via the probabilistic link prediction task, i.e., predicting the distribution of the existence of links between pairs of univariate time series. 
A \ac{cnn} was used to extract features from full-length time series, followed by an additive attention mechanism to infer the probability of the link existence between pairs of features using a Bernoulli distribution. 
They then uses a reparameterization trick based on the Gumbel sampling to obtain a sparse graph. 
The authors finally integrated the discrete graph learning into \ac{dcrnn} framework to obtain an end-to-end prediction framework.
Wu \etal \cite{mtgnn_2020} proposed \ac{mtgnn} framework consisting of the graph learning module in addition to the graph convolutional module and the temporal convolution module. 
The graph learning module learns to extract a sparse graph adjacency matrix using a sampling approach, node embeddings, and attention mechanism. We will summarize the model in further detail in Section \ref{sect:seagnn}. 
% \Tableref{tab:design_block} summarizes the design techniques for spatial and temporal blocks used in previous work.

The comprehensive survey and analysis of spatial-temporal forecasting models indicate that most of them do not explicitly consider the time component, which we believe is the key component in time series.
Also, it remains unclear how to effectively combine information extracted from the time component with the one extracted from temporal values via graph convolution and/or temporal convolution modules. 
Our work is therefore focused on developing conditional neural fields and a layer-wise gated fusion mechanism to address the two issues.

\section{Our Approach}
\label{sect:methods}
    After summarizing the problem formulation, we describe the detail of our proposed approach for integrating the time component into forecasting models. We then present two novel forecasting models, namely \ac{seacnn} and \ac{seagnn} which are respectively the integration of the time component into the inception forecasting model and the state-of-the-art graph-based \ac{mtgnn} forecasting model \cite{mtgnn_2020} using our proposed approach in order to demonstrate that the proposed approach is general and can be readily implemented in any other time series forecasting models.

%%////////////////////
\subsection{Problem Formulation}
\label{subsec:problem_statement}

Time series forecasting is defined as the forecasting of the values of the future $T_f$ timesteps given the data of the historical $T_h$ timesteps. 
We denote $\mX$ as the multivariate time series representing the traffic data, and $\mX^{(t)}$ as the multivariate time series at time $t$. 
Thus $\mX \in \sR^{L\times N}$ and $\mX^{(t)} \in \sR^N$, where $L$ and $N$ are the length of the multivariate time series and the number of the variables, i.e., sensors, respectively. 
Multivariate traffic forecasting can be formulated as an optimization problem of finding a function $f$ as follows:
\begin{equation}
\label{eq:prob_formulation}
    \begin{aligned}
        & \argmin_f  \gL(f([ \mX^{(t - T_h + 1)}, \mX^{(t - T_h + 2)}, \dots, \mX^{(t)} ], \\
        & \qquad [\mX^{(t + 1)}, \mX^{(t + 2)}, \dots, \mX^{(t + T_f)} ])),
    \end{aligned}
\end{equation}
where $\gL$ is the loss function, which could be one of the well-defined metrics such as \ac{mae}, \ac{rmse}, and \ac{mape}. 
If we represent the traffic network sensors as a weighted directed graph $\gG=\left(\gV, \gE, \mW\right)$, where $\gV$ is a set of nodes with $|\gV|=N$, $\gE$ is a set of edges and $\mW \in \sR^{N\times N}$ is a weighted adjacency matrix representing the node proximity, the optimization problem \Eqref{eq:prob_formulation} becomes:
\begin{equation}
\label{eq:prob_formulation_graph}
    \begin{aligned}
        & \argmin_f  \gL(f(\gG, [ \mX^{(t - T_h + 1)}, \mX^{(t - T_h + 2)}, \dots, \mX^{(t)} ], \\
        & \qquad [\mX^{(t + 1)}, \mX^{(t + 2)}, \dots, \mX^{(t + T_f)} ])).
    \end{aligned}
\end{equation}

%%////////////////////
\subsection{Integrating Time Component into Time Series Forecasting Models}
\label{sect:framework}
The approach consists of three main blocks: auxiliary time feature extraction, conditional neural field module, and layer-wise gated fusion.

\subsubsection{Auxiliary Feature Extraction}

\begin{figure}
    \centering
    \includegraphics[width=0.95\columnwidth]{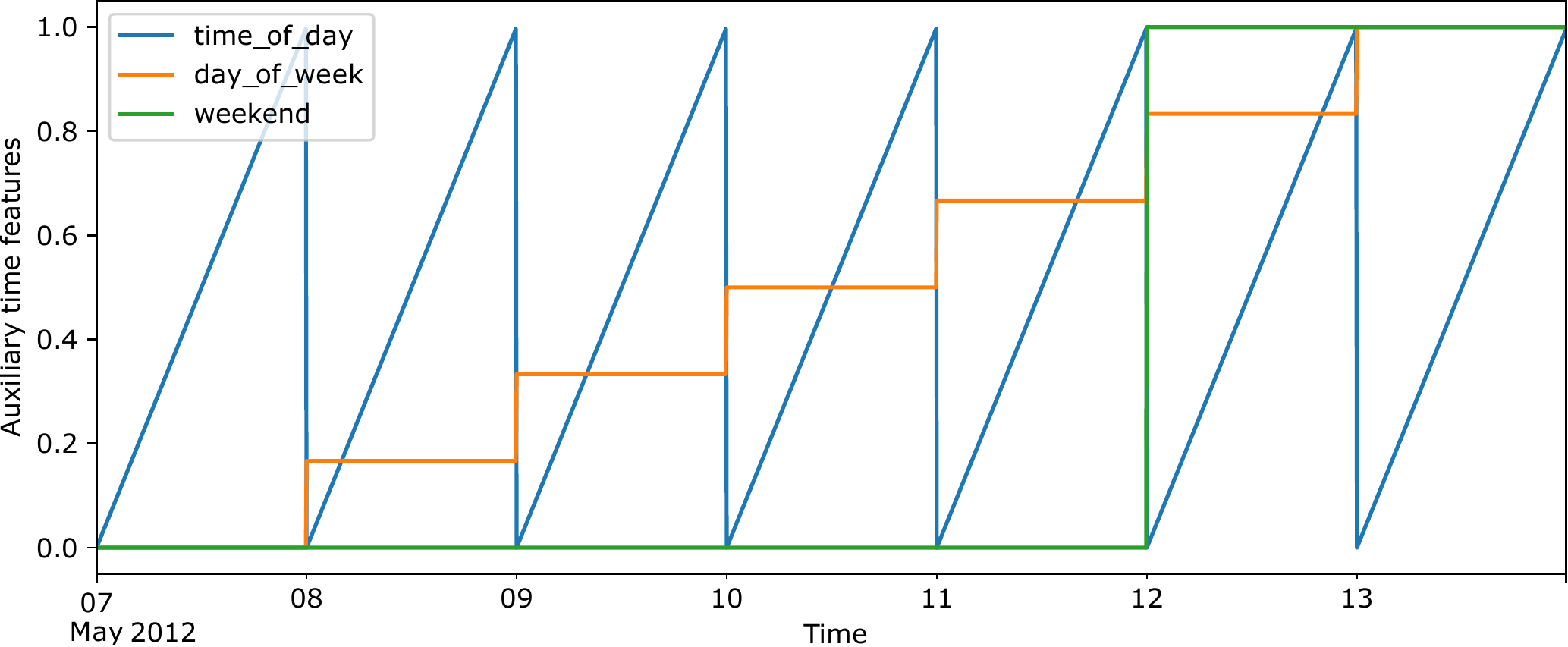}
    \caption{Auxiliary features extracted from the time component.}
    \label{fig:aux_features}
\end{figure}

The input to the neural field module should be numerical values, therefore auxiliary time features need to be extracted from the date time information of the time series. As can be seen from \Figref{fig:traffic_data}, time series contain repetitive patterns according to time, i.e., seasonality. The seasonality could happen daily and weekly, thus in order to incorporate those seasonality patterns, we need to extract relevant features such as time of day, day of the week, and weekend as illustrated in \Figref{fig:aux_features}. Furthermore, if the data contain sufficient long history, other features such as month and holiday could be considered.

%%%%%%%%%%%%%%%%%%%%%%%%%%%%%%%%%%%%%%%%%%%%%%%%%%%%%%%%%%
\subsubsection{Conditional Neural Field (CNF) Module}

\begin{figure}
    \centering
    \includegraphics[width=0.9\columnwidth]{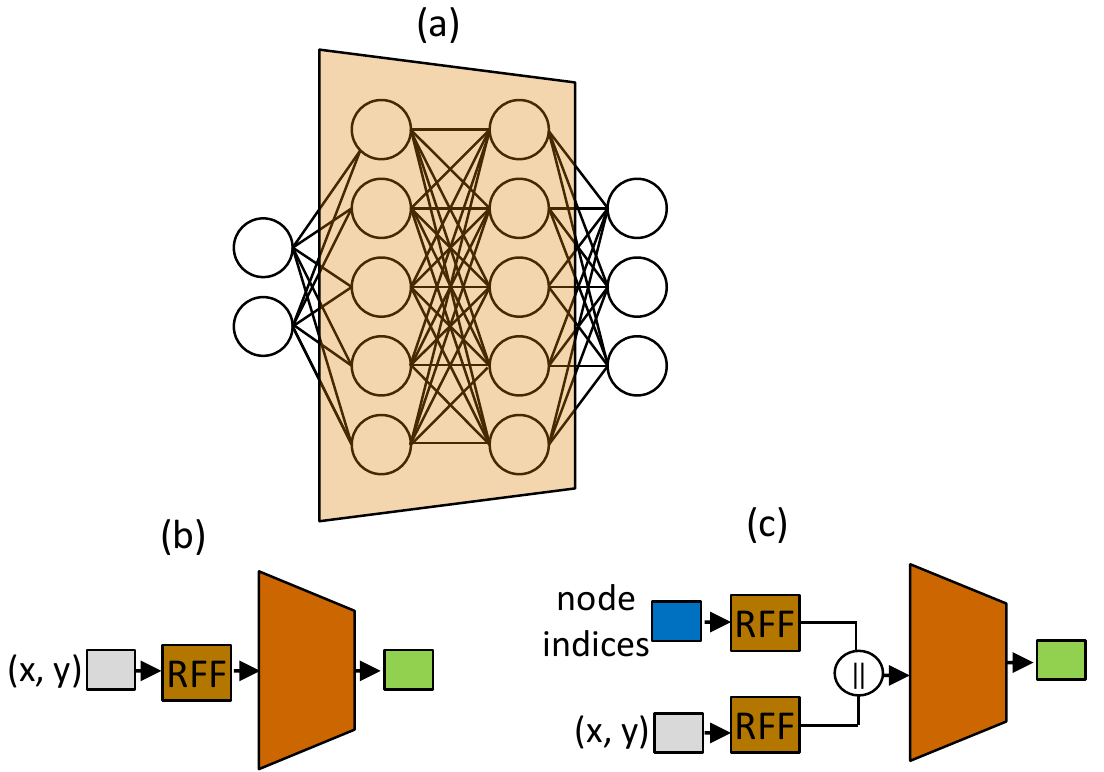}
    \caption{Neural field modules. Simple neural field module implemented with two-layer MLP network (a), neural field with Random Fourier features (RFF) module (b), and conditional neural field module with RFF (c).}
    \label{fig:neural_field}
\end{figure}

The main objective is to use the neural field module to learn the seasonality patterns that will be considered as global information of time series. As the neural field is the coordinate-based \ac{mlp}, we treat the auxiliary features of time series as coordinates. The neural field model is described as follows:
\begin{align}
    &\Phi(\vx) = F_n(\text{ReLU}(F_{n-1}(\dots (\text{ReLU}(F_1(\vx)))))),
\end{align}
where $\vx$ are auxiliary features, $n$ is the number of layers, $F_n$ is a feed-forward function of the form $Wx + b$. \Figref{fig:neural_field}a is an example of a two-layer neural field module.

\textbf{Expressivity}
General \ac{mlp}s are known to suffer from the spectral bias problem, which limits the reconstructed signals to the low-frequency spectrum and fails to learn the high-frequency components present in the data. Therefore, we adopted the \ac{rff} \cite{fourfeat} to overcome the problem. Both theoretically and experimentally, Tancik \etal \cite{fourfeat} showed that passing the input coordinates through a simple Fourier feature mapping allows the \ac{mlp} to overcome the spectral bias and succeed in learning the high-frequency signals:
\begin{align}
    \gamma(\vx) = [\cos(2\pi\mB \vx), \sin(2 \pi \mB \vx )] ^\top,
\end{align}
where $\vx \in \sR^d$ is the input coordinates and $\mB \in \sR^{m \times d}$ is sampled from a Gaussian distribution $\gN(0, \sigma^2)$ with the standard deviation of $\sigma$ that can be chosen experimentally based on the data. One can relate this mapping to the positional encoding \cite{vaswani2017}. However, RFF is well-grounded by the theory and outperformed the positional encoding quantitatively in the reconstruction tasks \cite{fourfeat}. In addition, we performed an experiment to justify the choice of \ac{rff} for our approach as presented in Appendix \ref{sect:appendix_a}.

\textbf{Conditional Neural Fields (CNF)}
A neural field can represent a time series either uni- or multivariate. In the traffic prediction literature, to exploit the spatial dependencies of multiple time series, we usually group individual time series acquired from each sensors (i.e., nodes) into a multivariate time series of a large number of variables, e.g.,  207 for METR-LA and 325 for PEMS-BAY dataset. Consequently, we design the neural field module towards a spatially-aware model which can produce different features not only for different timestamps but also for different sensors/nodes. A simple solution could be to implement $N$ neural field modules for $N$ sensors. However, this makes the model complexity multiplied by a factor of $N$, avoiding the scalability of the model. Therefore, we design a \ac{cnf} module, which allows varying a neural field by latent vector $\vz$. In our traffic prediction, as the set of sensors is finite, it is not necessary to sample $\vz$ from a prior distribution. Instead, we retrieve $\vz$ deterministically from the node indices. We also use the RFF to encode the node indices into the node positional embeddings. We then concatenate the input coordinates with the latent vectors of nodes to retrieve features per timestamp per node as illustrated in \Figref{fig:neural_field}c.
 
In summary, we implement the neural field as a coordinate-based \ac{mlp} $\Phi: \gD \times \sE \rightarrow \sR^{d}$, which maps from the domain of coordinates $\gD = [0, 1]^2$ and the set of node embeddings $\sE \subset \sR^{d_e}$ to the feature space $\sR^{d}$, where $d_e$ is the dimension of the node embeddings and $d$ is the dimension of the feature space. $\Phi$ is implemented with \ac{rff} to enhance the expressivity of the neural field. With the inputs of the timestamps and the node embeddings, $\Phi$ produces outputs as the global information. The global information will be combined with the local features that are extracted from relevant subsequences using autoregressive neural networks to produce the final predictions. We will discuss the global-local fusion mechanism in the next section.

%%%%%%%%%%%%%%%%%%%%%%%%%%%%%%%%%%%%%%%%%%%%%%%%%%%%%%%%%%%%%%%%%%%%%
\subsubsection{Layer-wise Gated Fusion (LGF)}

There exists a trade-off between global and local features. Global features take into account the daily and weekly seasonality of the time series, which are robust to noises. Local features take into account the dynamic characteristics of time series, which are able to capture the local variation within a short-range time period. Choosing the global features over the local ones may risk limiting the prediction model from being reactive to the dynamics of the time series including abnormal behaviors (e.g., outliers or anomalies), whereas choosing the local over global features makes the model vulnerable to noises. In order to balance between global and local features, we propose the use of \textit{gated fusion} as follows:
\begin{align}
    &\vz = \sigma((\mH_\text{local} \| \mH_\text{global}) \mW + \vb), \\
    &\mH = (1 - \vz) \odot \mH_\text{local} + \vz \odot \mH_\text{global},
\end{align}
where $\|$ and $\odot$ are the concatenation operator and the Hadamard product; $\mH_\text{local}$ and $\mH_\text{global}$ are the local and global extracted features; $\mW$ and $\vb$ are learnable weights and biases; $\sigma$ is the Sigmoid activation function. The architecture of this block can be depicted at the top right of \Figref{fig:seacnn_model}, which can be considered to be a soft and adaptive manner to combine the global and local features.

One question still remains: when should the two types of features be fused?
If the fusion of features is performed prior to feeding them into the autoregressive neural network model, the global features may exert an excessive influence on the gated fusion process.
If we fuse them after obtaining the local features from the autoregressive model, it has the risk that the global features have little interaction with individual modules of the autoregressive model such as temporal convolution or graph convolution modules. 
Therefore, we  propose the \ac{lgf} module to fuse the features within every layer (i.e., module) of the autoregressive network, as shown in \Figref{fig:seacnn_model}.

%%%%%%%%%%%%%%%%%%%%%%%%%%%%%%%%%%%%%%%%%%%%%%%%%%%%%%%%%%%%%%%%%%%%%%%
\subsection{\acrfull{seacnn}}
\label{sect:seacnn}
\begin{figure}[t]
    \centering
    \includegraphics[width=\columnwidth]{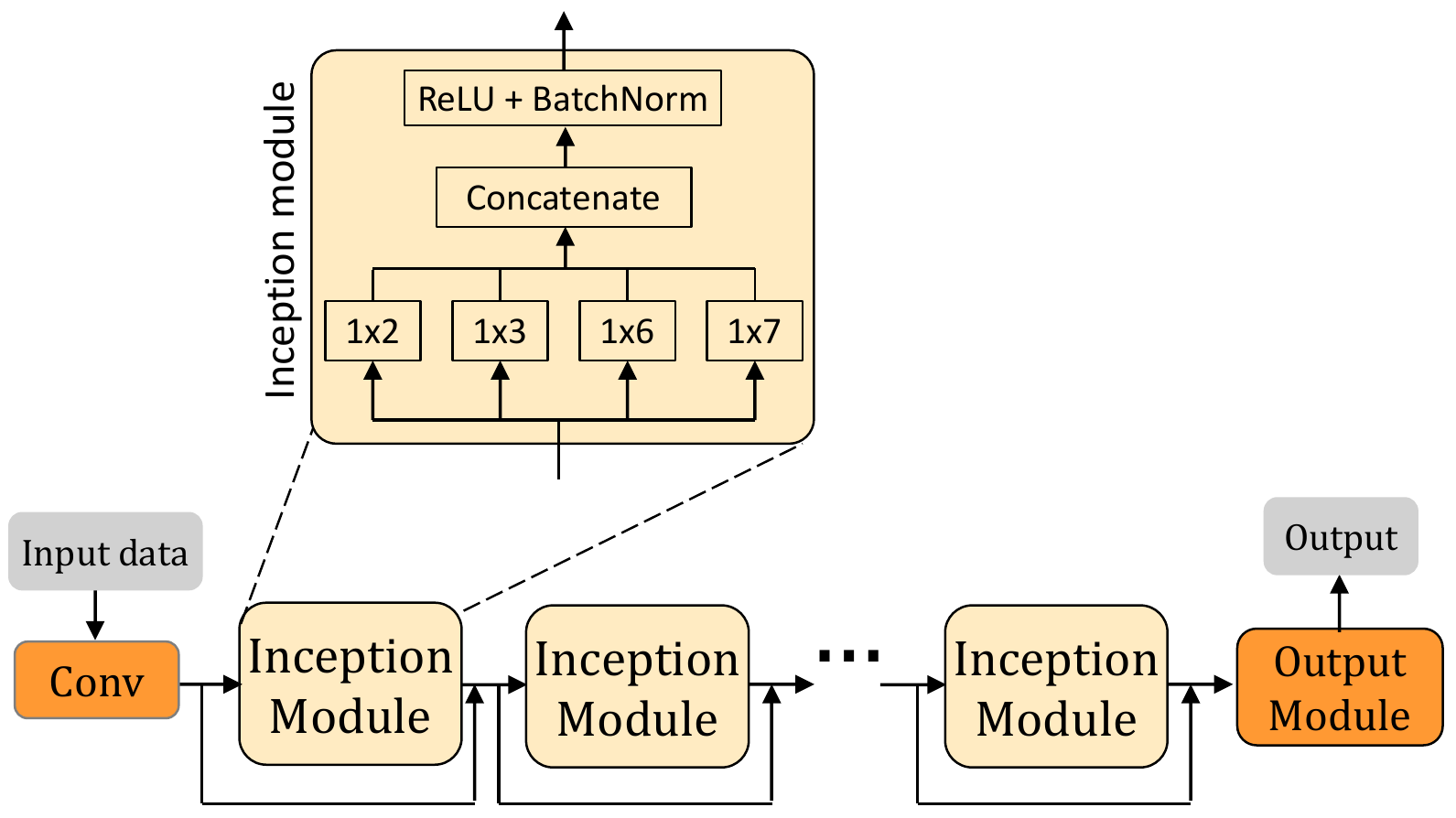}
    \caption{The architecture of the inception baseline forecasting model.}
    \label{fig:inception_model}
\end{figure}

In order to demonstrate the effectiveness of the proposed approach, we first implement an inception baseline forecasting model as shown in \Figref{fig:inception_model}. The model consists of a $1\times1$ standard convolution layer, $N$ inception modules, and an output module. The $1\times1$ standard convolution layer is used to transform the channel dimension of the inputs to the channel dimension of the inception module. As shown in \Figref{fig:inception_model}, the inception module is composed of the concatenation of 4 convolutions with 4 different kernel sizes in order to explore temporal patterns with different frequencies, followed by a ReLU activation and BatchNorm regularization. We also employ a residual connection for each inception module to improve the robustness of the model against the problem of gradient vanishing. Finally, the output module which comprises two $1\times1$ convolutions with a ReLU activation in between is used to mainly convert the channel dimension of the inception module to the desired output dimensions.

\begin{figure}[t]
    \centering
    \includegraphics[width=\columnwidth]{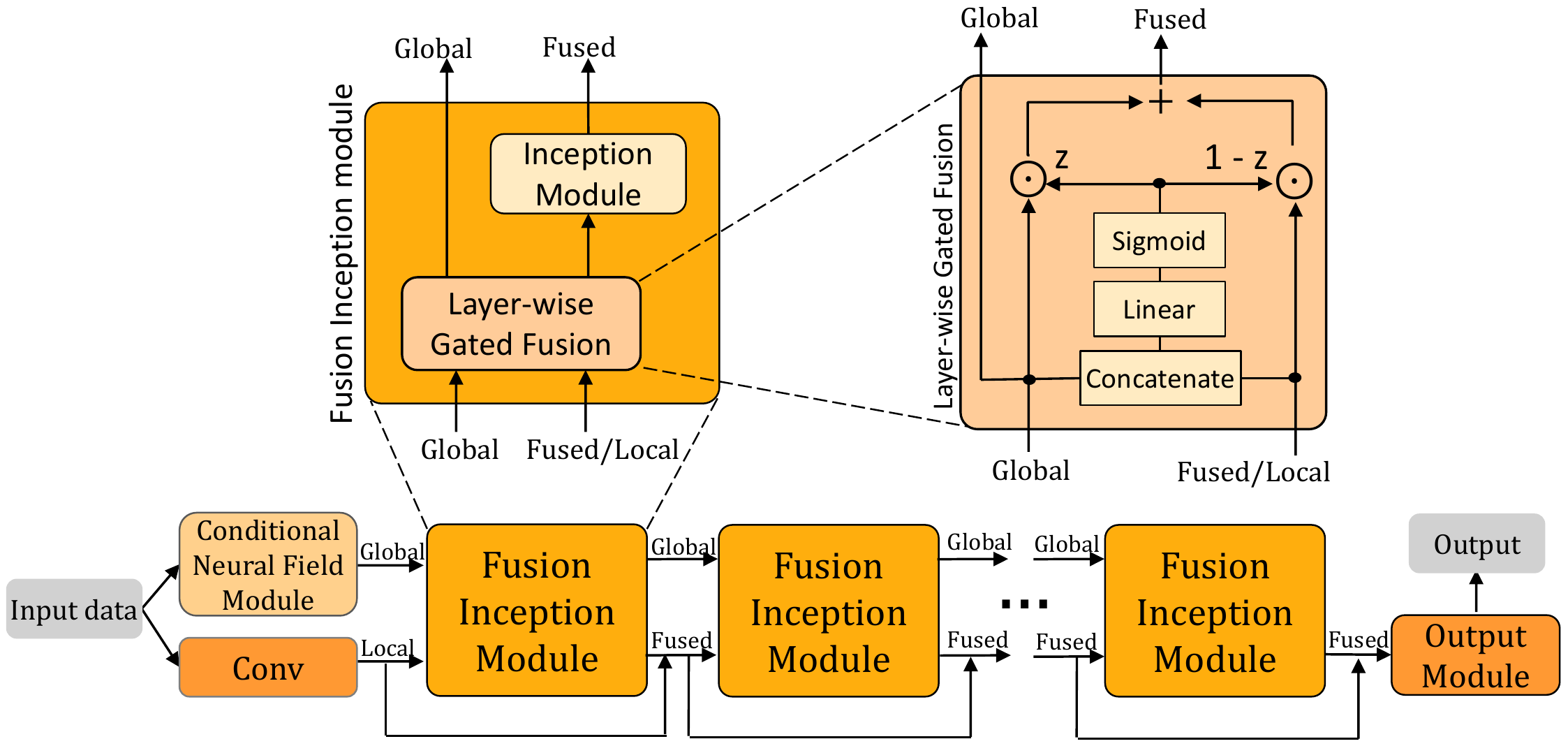}
    \caption{The architecture of the \ac{seacnn} model.}
    \label{fig:seacnn_model}
\end{figure}

\Figref{fig:seacnn_model} depicts the architecture of the \ac{seacnn} forecasting model which is the integration of the time component using the approach described in Section \ref{sect:framework} into the inception forecasting model described above. The main objective of the integration approach is to help the model aware of the seasonality patterns present in the data through the introduction of the \ac{cnf} module and the \ac{lgf} module. As shown in \Figref{fig:seacnn_model}, the time component and node indices of the input data are fed into the \ac{cnf} to extract the seasonality-aware information called global information which will be fed into each of all Fusion Inception modules. The Fusion Inception module is responsible for fusing the local/fused information extracted by the Conv/Inception module and the global information in order to output the fused information for the subsequent modules. The Conv layer and Output module are similar to those described in the inception baseline model. Therefore, the integration of the time component into the inception baseline model is quite straightforward, we then describe how the integration can be done with the state-of-the-art graph-based \ac{mtgnn} \cite{mtgnn_2020} forecasting model.

%%%%%%%%%%%%%%%%%%%%%%%%%%%%%%%%%%%%%%%%%%%%%%%%%%%%%%%%%%%%%%%%%%%%%%%%%%
\subsection{\acrfull{seagnn}}
\label{sect:seagnn}

\begin{figure*}
    \centering
    \includegraphics[width=0.8\textwidth]{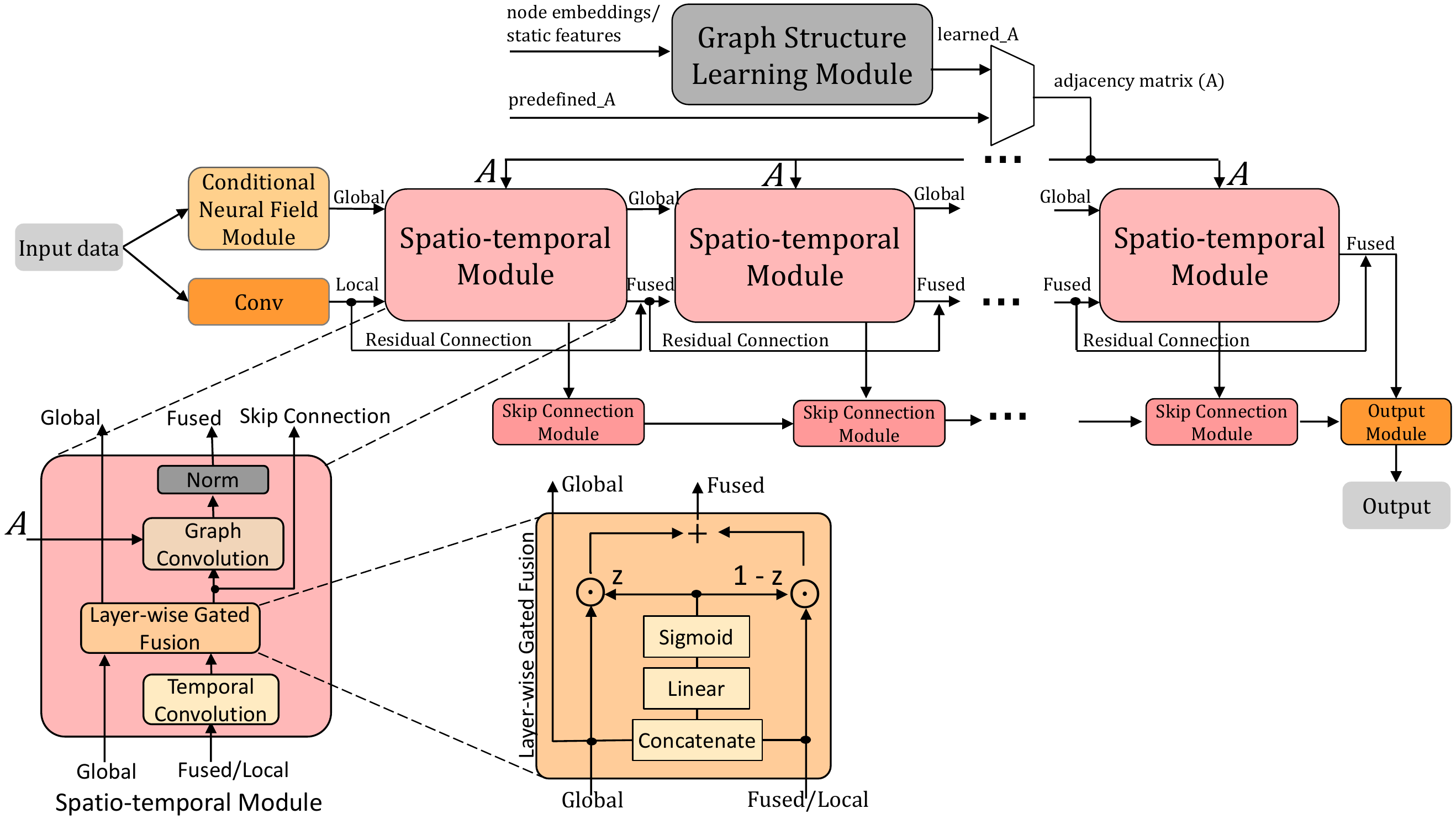}
    \caption{The architecture of the novel \ac{seagnn} model.}
    \label{fig:seagnn_model}
\end{figure*}

Wu \etal \cite{mtgnn_2020} proposed \ac{mtgnn} algorithm mainly consisting of the graph structure learning module, the graph convolution module, and the temporal convolution module. The graph learning module learns to extract a sparse graph adjacency matrix using a sampling approach, node embeddings, and attention mechanism. Given the extracted sparse adjacency matrix, the graph convolution module comprises two mix-hop propagation layers to handle information flow over spatially dependent nodes. The mix-hop propagation layer which contains the information propagation and the information selection steps is designed for directed graphs and avoids the over-smoothing problem. Finally, the temporal convolution consists of two dilated inception convolution layers, offering two advantages: (1) be able to capture temporal patterns with multiple frequencies, and (2) be able to process very long sequences thanks to its long receptive field coming from the dilated convolutions. In addition, the skip connection layers are $1\times L_i$ standard convolutions, where $L_i$ is the sequence length of the inputs to the $i^{th}$ skip connection layers in order to standardize the information. Together with the residual connection, they both help the model be more robust and avoid the vanishing gradient problem.

The integration of the time component into the \ac{mtgnn} model results in a novel forecasting \ac{seagnn} model as illustrated in \Figref{fig:seagnn_model}. The \ac{cnf} module, Conv layer, and Output module play similar roles as those described above in the \ac{seacnn} model. We introduce the spatio-temporal module which comprises temporal convolution, graph convolution, and \ac{lgf} module together with BatchNorm regularization layer. The temporal and graph convolution modules are similar to that of the MTGNN model. The \ac{lgf} module is used to fuse the global information extracted from the \ac{cnf} module and the local information extracted by the temporal module.

\section{Experiments}
\label{sect:experiments}

\subsection{Datasets, Evaluation Metrics and Experimental Settings}
\label{subsect:datasets_evaluation_metrics}

\begin{table*}[t]
\centering
\caption{Summary of datasets used in our experiments}
\label{tab:exp_dataset}
\begin{tabular}{l|cc|cc}
\hline
Tasks       & \multicolumn{2}{c|}{Speed prediction} & \multicolumn{2}{c}{Network traffic prediction} \\ \hline
Name        & METR-LA            & PeMS-BAY         & MI-Call                & MI-SMS                \\
Region      & Los Angeles        & Bay Area         & Milano                 & Milano                \\
Start date  & 03/01/2012         & 01/01/2017       & 10/31/2013             & 10/31/2013            \\
End date    & 06/27/2012         & 06/30/2017       & 01/01/2014             & 01/01/2014            \\
\# Days     & 119                & 181              & 62                     & 62                    \\
\# Nodes    & 207                & 325              & 207                    & 207                   \\
Granularity & 5 minutes          & 5 minutes        & 10 minutes             & 10 minutes            \\
Features    & Speed              & Speed            & Vol. of incoming calls  & Vol. of SMS received  \\ \hline
\end{tabular}
\end{table*}

The proposed methods are evaluated and compared with existing methods using two real-world road traffic open datasets, METR-LA and PEMS-BAY, and two cellular network traffic datasets \cite{barlacchi2015multi}, as summarized in Table \ref{tab:exp_dataset}.

METR-LA contains traffic information collected from loop detectors in the highway of Los Angeles County \cite{jagadish2014big}. As previous work \cite{dcrnn_2018}, we select 207 sensors and use 4 months of data ranging from March 1st, 2012 to June 30th, 2012 for the experiments. PEMS-BAYS is collected by California Transportation Agencies Performance Measurement System (PeMS). We select 325 sensors in the Bay Area and use 6 months of data ranging from Jan 1st, 2017 to June 30th, 2017 for the experiments. 
\begin{figure}
    \centering
    %\includesvg[inkscapelatex=false,width=\columnwidth]{figures/finetune_nef.svg}
    \includegraphics[width=\columnwidth]{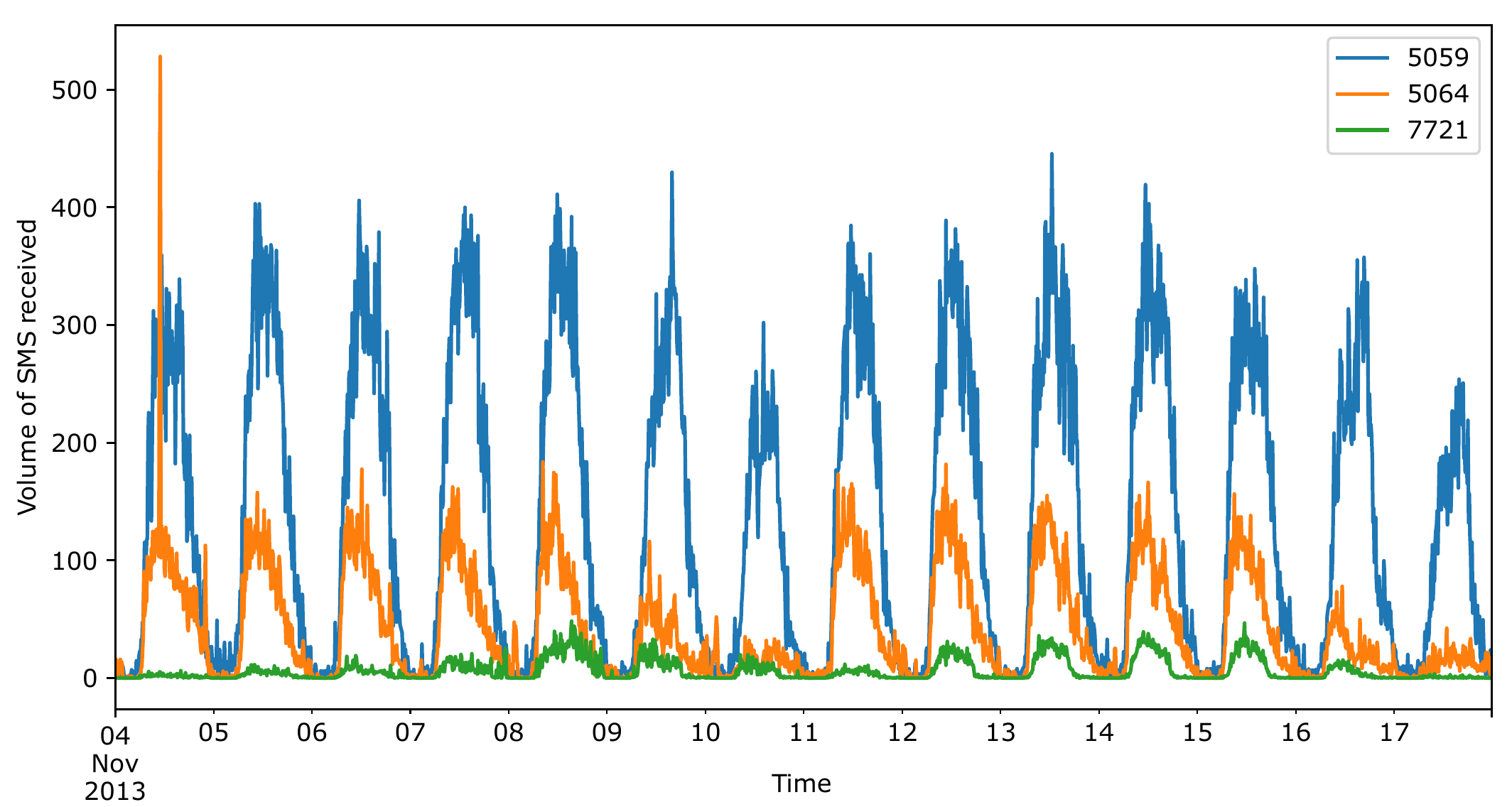}
    \caption{The volumes of SMS received of three network base stations in the city of Milan.}
    \label{fig:milan_sms}
\end{figure}

The cellular traffic datasets come from the ``Telecom Italia Big Data Challenge" \cite{barlacchi2015multi}. The data consists of time series of aggregated cell phone traffic including short message service (SMS) and call service received by users within the specific area over the city of Milan. The original datasets contain traffic data collected for 10000 base stations, i.e., cells, but we selected the most active cells of each day using information about cells to province and province to cells present in the data, resulting in 207 cells. Finally, we have 2 sub-datasets, named as MI-Call, MI-SMS corresponding to the volumes of incoming calls and SMS received for Milan city. \Figref{fig:milan_sms} presents the time series of the volumes of SMS received in 2 weeks from three cells for Milan city. The figure highlights the fact that the data are in different ranges of values, very noisy, and have strong daily and weekly seasonality.

For road traffic datasets, we split them into three parts in ascending time order with 70\% for training, 10\% for validation, and the remaining 20\% for testing as in previous work \cite{dcrnn_2018, mtgnn_2020}. For network traffic datasets, as the data length is shorter, we, therefore, split them into 80\% for training, 10\% for validation, and 10\% for testing.

For road traffic forecasting, we use three commonly used evaluation metrics, (1) \ac{mae}, (2) \ac{mape}, and (3) \ac{rmse}, which are defined as follows:
\begin{align}
    \text{MAE}=\frac{1}{N}\sum_{i=1}^{N} \left | \hat{y}_i -y_i \right |, \\
    \text{MAPE}=\frac{1}{N}\sum_{i=1}^{N} \left | \frac{\hat{y}_i -y_i }{y_i}\right |, \\
    \text{RMSE}=\sqrt{\frac{1}{N}\sum_{i=1}^{N}(\hat{y}_i-y_i)^2},
\end{align}
where $y_i$ and $\hat{y_i}$ are the measured and predicted values, $N$ is the length of the data under evaluation. 
The missing values are not taken into account while calculating these metrics. 
For network traffic data, because the values of many time series, i.e., cells, are close to zero as illustrated in Fig. \ref{fig:milan_sms} we replace MAPE metric by the \ac{smape} defined as follows:
\begin{align}
    \text{sMAPE}=\frac{1}{N}\sum_{i=1}^{N} \left | \frac{\hat{y}_i - y_i }{\hat{y}_i + y_i}\right |.
\end{align}

We obtained the final results on the testing data using two training schemes: (1) we train the model for 50 epochs and (2) we train the model for 100 epochs with a curriculum learning \cite{mtgnn_2020}, which is the training scheme allowing the model to be trained from easy tasks to more difficult tasks. In the context of time series forecasting, curriculum learning enables the model to learn to predict from the short prediction horizons to the longer ones. In practice, we increase the prediction horizon by 1 every 2500 training iterations until it reaches the 12-step prediction horizon, corresponding to the 60-minute horizon. The training loss is computed as follows:
\begin{align}
    \gL_\text{training} = \sum_{i=0}^N \left | \hat{\mY} _{i:i+p} - \mY _{i:i+p}\right |,
\end{align}
where $p$ is the training prediction horizon that starts from 1 and increases by 1 in every 2500 iterations until it reaches 12.

\subsection{Results on Road Traffic Open Datasets}
\label{subsect:road_traffic_results}
\Tableref{tab:road_results} presents the results of the inception baseline, \ac{seacnn}, \ac{mtgnn} and \ac{seagnn} forecasting models together with existing ARIMA statistical model, LSTM deep learning model, and \ac{dcrnn} graph-based model on the two road traffic open datasets. Note that the results of ARIMA, LSTM, and \ac{dcrnn} models were taken from previous works \cite{dcrnn_2018, mtgnn_2020} because we used exactly the same evaluation protocol. For our four implemented models, we performed the experiments 5 times by varying random seed numbers together with the curriculum learning scheme and computed the mean and standard deviation of the results.

As can be seen from the table, the inception model proves to be a very strong baseline forecasting model that outperforms the ARIMA and LSTM models by a large margin for all prediction horizons on the two datasets. It also approaches the performance of the graph-based \ac{dcrnn} model, especially on the PEMS-BAY dataset. \Tableref{tab:road_results} also shows that the \ac{seacnn} model provides better results than the inception model across all prediction horizons on both datasets which proves the effectiveness of the integration approach of the time component.

The results of the \ac{mtgnn} model with the curriculum learning scheme are slightly better than the results reported in the paper \cite{mtgnn_2020}, especially on the METR-LA dataset, e. g., MAE of 2.654 vs 2.69 for 15min prediction horizon. In addition, \Tableref{tab:road_results} indicates that \ac{seagnn} provides better performance than \ac{mtgnn} model across all prediction horizons on both datasets. To the best of my knowledge, with the MAE of 2.623 for 15min prediction horizon on METR-LA, SEAGNN achieves the best results compared to all other state-of-the-art forecasting models reported for this dataset. 

\begin{table*}[]
\centering
\caption{Results on Road Traffic Open Datasets}
\label{tab:road_results}
\begin{tabular}{c|c|c|ccccccc}
\hline
\multirow{2}{*}{Datasets} & \multirow{2}{*}{T}     & \multirow{2}{*}{Metrics} & \multirow{2}{*}{ARIMA} & \multirow{2}{*}{LSTM} & \multirow{2}{*}{DCRNN} & \multirow{2}{*}{Inception}          & \multicolumn{1}{c}{\multirow{2}{*}{SEACNN}} & \multirow{2}{*}{MTGNN}              & \multirow{2}{*}{SEAGNN}            \\
                          &                        &                         &                        &                       &                        &                                     & \multicolumn{1}{c}{}                        &                                     &                                    \\ \hline
\multirow{9}{*}{METR-LA}  & \multirow{3}{*}{15min} & MAE                     & 3.99                   & 3.44                  & 2.77                   & 2.944 \textpm \  0.000      & 2.856 \textpm \  0.000               & 2.654 \textpm \  0.001      & 2.623 \textpm \  0.000     \\
                          &                        & RMSE                    & 8.21                   & 6.3                   & 5.38                   & 5.773 \textpm \  0.002      & 5.666 \textpm \  0.010              & 5.116 \textpm \  0.006      & 5.060 \textpm \  0.000      \\
                          &                        & MAPE (\%)                   & 9.60                 & 9.60                & 7.30                 & 7.896 \textpm \  0.019  & 7.778 \textpm \  0.034          & 6.853 \textpm \  0.018  & 6.767 \textpm \  0.003 \\ \cline{2-10} 
                          & \multirow{3}{*}{30min} & MAE                     & 5.15                   & 3.77                  & 3.15                   & 3.498 \textpm \  0.001      & 3.257 \textpm \  0.000              & 3.015 \textpm \  0.001      & 2.967 \textpm \  0.000      \\
                          &                        & RMSE                    & 10.45                  & 7.23                  & 6.45                   & 7.052 \textpm \  0.001      & 6.747 \textpm \  0.012              & 6.092 \textpm \  0.002      & 6.029 \textpm \  0.001     \\
                          &                        & MAPE (\%)                  & 12.70                & 10.90               & 8.80                 & 10.094 \textpm \  0.033 & 9.406 \textpm \  0.031          & 8.231 \textpm \  0.029  & 8.075 \textpm \  0.007 \\ \cline{2-10} 
                          & \multirow{3}{*}{60min} & MAE                     & 6.9                    & 4.37                  & 3.6                    & 4.185 \textpm \  0.001      & 3.679 \textpm \  0.002              & 3.432 \textpm \  0.009      & 3.359 \textpm \  0.002     \\
                          &                        & RMSE                    & 13.23                  & 8.69                  & 7.59                   & 8.454 \textpm \  0.003      & 7.695 \textpm \  0.010              & 7.071 \textpm \  0.010      & 6.990 \textpm \  0.014     \\
                          &                        & MAPE (\%)                    & 17.40                & 13.20               & 10.50                & 12.974 \textpm \  0.055 & 11.139 \textpm \  0.038         & 10.008 \textpm \  0.025 & 9.725 \textpm \  0.005 \\ \hline
\multirow{9}{*}{PEMS-BAY} & \multirow{3}{*}{15min} & MAE                     & 1.62                   & 2.05                  & 1.38                   & 1.380 \textpm \  0.000      & 1.343 \textpm \  0.001              & 1.326 \textpm \  0.002      & 1.312 \textpm \  0.001     \\
                          &                        & RMSE                    & 3.3                    & 4.19                  & 2.95                   & 2.955 \textpm \  0.002      & 2.847 \textpm \  0.002              & 2.793 \textpm \  0.002      & 2.763 \textpm \  0.000     \\
                          &                        & MAPE (\%)                    & 3.50                 & 4.80                & 2.90                 & 2.895 \textpm \  0.001  & 2.840 \textpm \  0.001          & 2.797 \textpm \  0.012  & 2.780 \textpm \  0.005 \\ \cline{2-10} 
                          & \multirow{3}{*}{30min} & MAE                     & 2.33                   & 2.2                   & 1.74                   & 1.797 \textpm \  0.001      & 1.690 \textpm \  0.003              & 1.647 \textpm \  0.004      & 1.619 \textpm \  0.001     \\
                          &                        & RMSE                    & 4.76                   & 4.55                  & 3.97                   & 4.081 \textpm \  0.007      & 3.844 \textpm \  0.002              & 3.745 \textpm \  0.013      & 3.690 \textpm \  0.003     \\
                          &                        & MAPE (\%)                    & 5.40                 & 5.20                & 3.90                 & 4.120 \textpm \  0.007  & 3.854 \textpm \  0.001          & 3.713 \textpm \  0.017  & 3.651 \textpm \  0.004 \\ \cline{2-10} 
                          & \multirow{3}{*}{60min} & MAE                     & 3.38                   & 2.37                  & 2.07                   & 2.258 \textpm \  0.003      & 2.018 \textpm \  0.006              & 1.949 \textpm \  0.002      & 1.899 \textpm \  0.001     \\
                          &                        & RMSE                    & 6.5                    & 4.96                  & 4.74                   & 5.123 \textpm \  0.015      & 4.596 \textpm \  0.007              & 4.476 \textpm \  0.004      & 4.354 \textpm \  0.000     \\
                          &                        & MAPE (\%)                    & 8.30                 & 5.70                & 4.90                 & 5.566 \textpm \  0.017  & 4.783 \textpm \  0.009          & 4.612 \textpm \  0.002  & 4.452 \textpm \  0.018 \\ \hline
\end{tabular}
\end{table*}

\begin{figure}
    \centering
    %\includesvg[inkscapelatex=false,width=\columnwidth]{figures/val_losses.svg}
    \includegraphics[width=0.9\columnwidth]{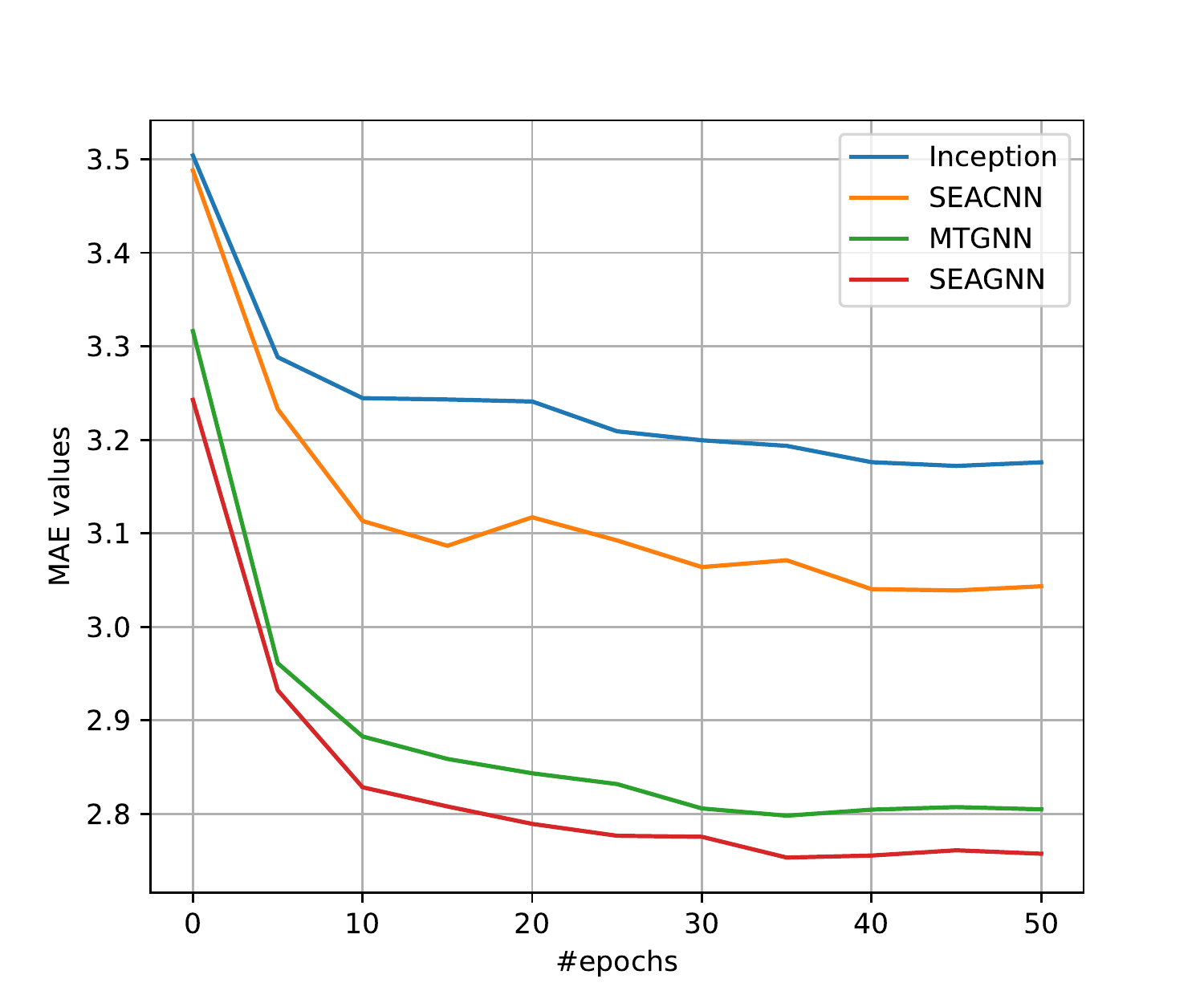}
    \caption{The evolution of MAE metrics on the validation data of METR-LA dataset for Inception, \acrshort{seacnn}, \acrshort{mtgnn} and \acrshort{seagnn} methods.}
    \label{fig:val_mae}
\end{figure}

\Figref{fig:val_mae} shows the evolution of the \ac{mae} metrics on the METR-LA validation data during the training process of the inception, \ac{seacnn}, \ac{mtgnn} and \ac{seagnn} forecasting models. The figure shows that \ac{seacnn} and \ac{seagnn} consistently provide better \ac{mae} metric and converge faster than their inception and \ac{mtgnn} counterparts, and that \ac{seagnn} is the most effective model.

\begin{figure}
    \centering
    %\includesvg[inkscapelatex=false,width=\columnwidth]{figures/metr_la_163_full_.svg}
    \includegraphics[width=\columnwidth]{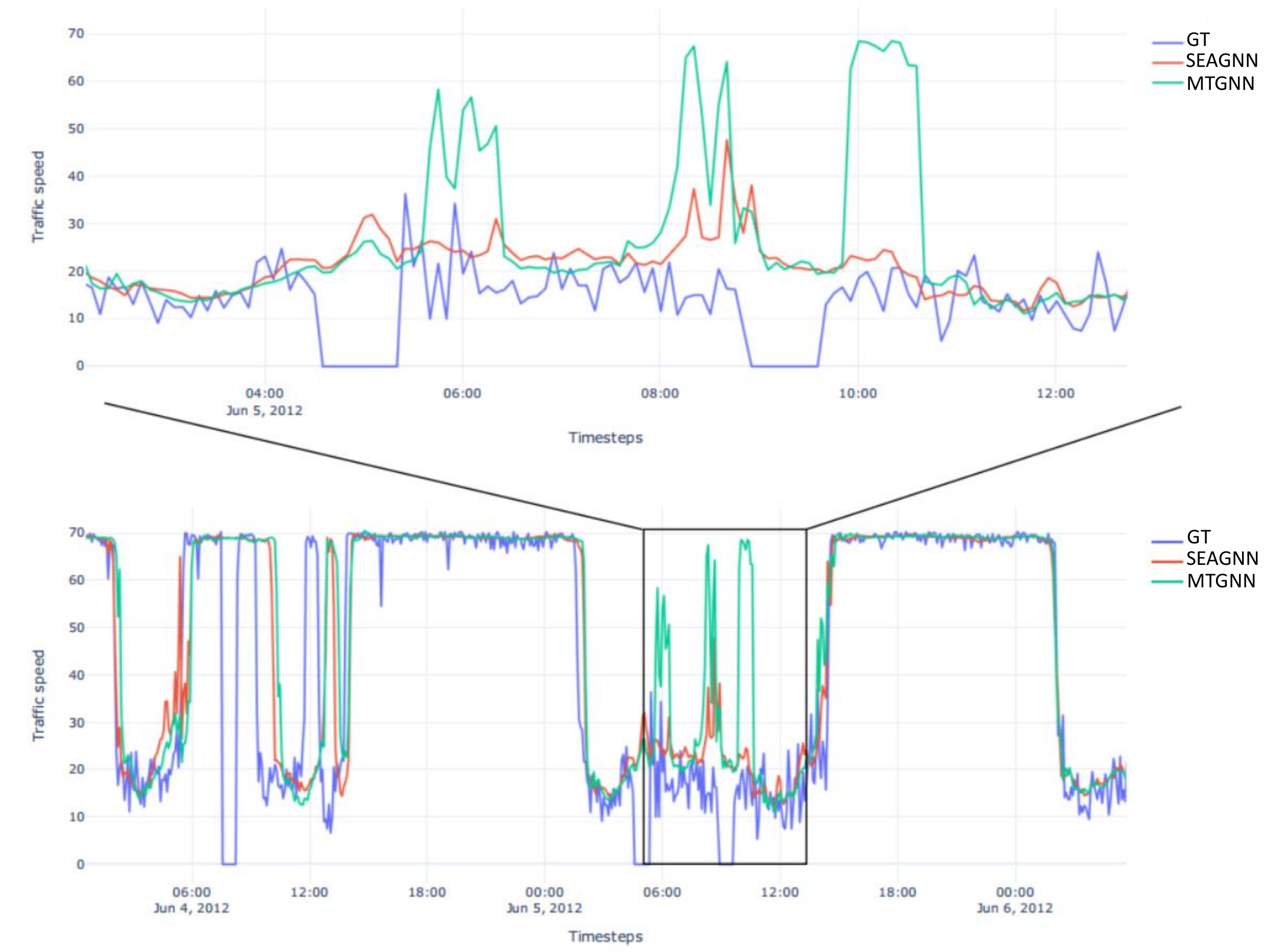}
    \caption{Prediction results of the 60-minute horizon on node 163 of the METR-LA dataset.}
    \label{fig:metr_result_163}
\end{figure}

\begin{figure}
    \centering
    %\includesvg[inkscapelatex=false,width=\columnwidth]{figures/metr_la_163_full_.svg}
    \includegraphics[width=\columnwidth]{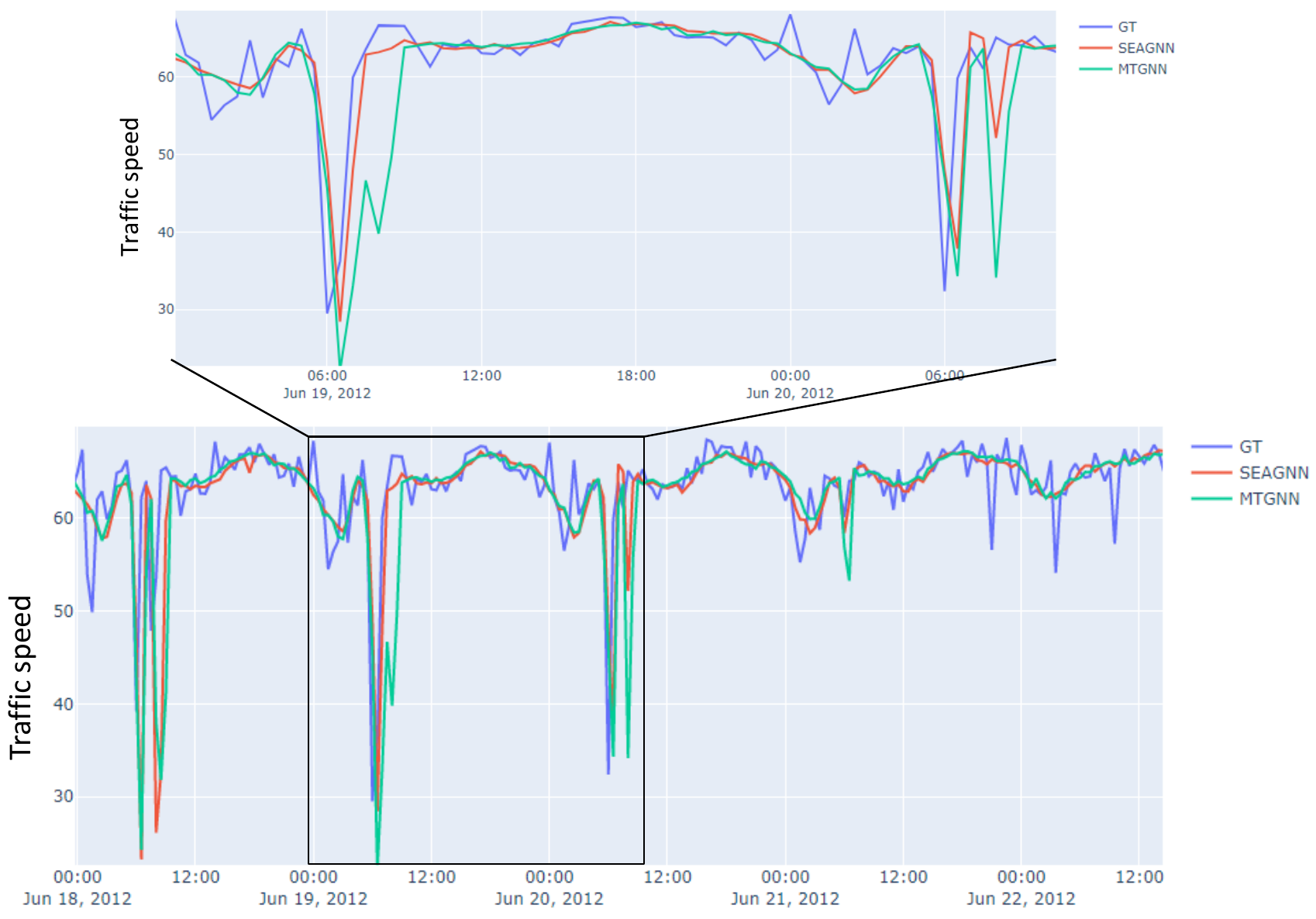}
    \caption{Prediction results of the 60-minute horizon on node 185 of the METR-LA dataset.}
    \label{fig:metr_result_185}
\end{figure}

The prediction results are examined qualitatively as shown in \Figref{fig:metr_result_185} and \Figref{fig:metr_result_163}. Compared to \ac{mtgnn} results, \ac{seagnn} provides prediction results that are more robust to missing values and approximate better the pattern of the ground truth.

It can be concluded that the integration approach is not only effective in the convolution-based model but also in the graph-based forecasting model.

\textbf{Ablation Study:}

We perform an ablation study where we evaluate the impact of the \ac{rff}, the LGF, and the aggregation modes on the final predictions on the validation split of the \textbf{METR-LA} dataset of the \ac{seagnn} model. We name our method with different variants as follows:
\begin{itemize}
    \item \textbf{w/o RFF}: we replace the RFF encoding with a learnable linear layer.
    \item \textbf{w/o LGF}: we remove the LGF and only fuse the global and local features before the first spatio-temporal module by addition.
    \item \textbf{Aggregation mode}: we replace the gated fusion with other different aggregation methods such as addition, multiplication, and concatenation.
\end{itemize}
We run the experiments 5 times with different random seeds and observe the evaluation metrics for all prediction horizons (i.e., 1 - 12). \Tableref{tab:rff_abla} summarizes statistics of the results. As can be seen from the table, all choices of the model have created negative impacts on the performance of the forecasting model in which SEAGNN provides the best results.

\begin{table}
\caption{Results of ablation study on the validation data for all prediction horizons from 1 (5 minutes) - 12 (60 minutes).}
\label{tab:rff_abla}
\resizebox{\columnwidth}{!}{%
\begin{tabular}{@{}ccccccc@{}}
\toprule
\multirow{2}{*}{Metrics} & \multirow{2}{*}{SEAGNN} & \multirow{2}{*}{w/o RFF} & \multirow{2}{*}{w/o LGF} & \multicolumn{3}{c}{Aggregation Modes}                                  \\ \cmidrule(l){5-7} 
                        &                       &                          &                          & Addition              & Multiplication        & Concatenation         \\ \midrule
MAE                     & \textbf{2.7369 \textpm \ 0.0041} & 2.7725 \textpm \ 0.0030    & 2.7515 \textpm \ 0.0090    & 2.8149 \textpm \ 0.0103 & 2.8452 \textpm \ 0.0090 & 2.8237 \textpm \ 0.0178 \\
RMSE                    & \textbf{5.6894 \textpm \ 0.0129} & 5.7445 \textpm \ 0.0160    & 5.7242 \textpm \ 0.0345    & 5.8289 \textpm \ 0.0244 & 5.9020 \textpm \ 0.0244 & 5.8365 \textpm \ 0.0345 \\
MAPE                    & \textbf{0.0764 \textpm \ 0.0003} & 0.0772 \textpm \ 0.0003    & 0.0770 \textpm \ 0.0010    & 0.0787 \textpm \ 0.0004 & 0.0793 \textpm \ 0.0009 & 0.0793 \textpm \ 0.0005 \\ \bottomrule
\end{tabular}%
}
\end{table}

\subsection{Results on Network Traffic Datasets}

\begin{table*}[]
\centering
\caption{Results on the Cellular Network Traffic Data of Milano 
}
\label{tab:network_results}
\begin{tabular}{c|c|c|cccc}
\hline
\multirow{2}{*}{Datasets} & \multirow{2}{*}{T}     & \multirow{2}{*}{Metrics} & \multirow{2}{*}{Inception}          & \multirow{2}{*}{SEACNN}             & \multirow{2}{*}{MTGNN}              & \multirow{2}{*}{SEAGNN}             \\
                          &                        &                         &                                     &                                     &                                     &                                     \\ \hline
\multirow{9}{*}{MI-Call} & \multirow{3}{*}{15min} & MAE                     & 1.562 \textpm \ 0.001      & 1.581 \textpm \ 0.005      & 1.482 \textpm \ 0.010      & 1.457 \textpm \ 0.004      \\
                          &                        & RMSE                    & 2.595 \textpm \ 0.008      & 2.588 \textpm \ 0.013      & 2.480 \textpm \ 0.019      & 2.397 \textpm \ 0.013      \\
                          &                        & sMAPE (\%)                   & 8.372 \textpm \ 0.025  & 8.914 \textpm \ 0.102  & 7.723 \textpm \ 0.068  & 7.817 \textpm \ 0.171  \\ \cline{2-7} 
                          & \multirow{3}{*}{30min} & MAE                     & 2.923 \textpm \ 0.003      & 2.903 \textpm \ 0.005      & 2.566 \textpm \ 0.004      & 2.500 \textpm \ 0.019      \\
                          &                        & RMSE                    & 5.036 \textpm \ 0.007      & 4.886 \textpm \ 0.029      & 4.478 \textpm \ 0.011      & 4.262 \textpm \ 0.053      \\
                          &                        & sMAPE (\%)                   & 13.273 \textpm \ 0.002 & 13.636 \textpm \ 0.053 & 11.118 \textpm \ 0.033 & 11.198 \textpm \ 0.127 \\ \cline{2-7} 
                          & \multirow{3}{*}{60min} & MAE                     & 4.104 \textpm \ 0.003      & 3.815 \textpm \ 0.004      & 3.404 \textpm \ 0.003      & 3.338 \textpm \ 0.019      \\
                          &                        & RMSE                    & 7.507 \textpm \ 0.018      & 6.753 \textpm \ 0.033      & 6.210 \textpm \ 0.037      & 5.983 \textpm \ 0.069      \\
                          &                        & sMAPE (\%)                   & 16.609 \textpm \ 0.017 & 16.217 \textpm \ 0.016 & 13.345 \textpm \ 0.034 & 13.243 \textpm \ 0.081 \\ \hline
\multirow{9}{*}{MI-SMS}  & \multirow{3}{*}{15min} & MAE                     & 2.679 \textpm \ 0.003      & 2.676 \textpm \ 0.012      & 2.632 \textpm \ 0.003      & 2.570 \textpm \ 0.015      \\
                          &                        & RMSE                    & 4.263 \textpm \ 0.005      & 4.224 \textpm \ 0.037      & 4.189 \textpm \ 0.012      & 4.054 \textpm \ 0.038      \\
                          &                        & sMAPE (\%)                   & 7.080 \textpm \ 0.012  & 7.138 \textpm \ 0.072  & 7.058 \textpm \ 0.034  & 6.847 \textpm \ 0.007  \\ \cline{2-7} 
                          & \multirow{3}{*}{30min} & MAE                     & 4.859 \textpm \ 0.000      & 4.786 \textpm \ 0.030      & 4.590 \textpm \ 0.014      & 4.458 \textpm \ 0.023      \\
                          &                        & RMSE                    & 7.900 \textpm \ 0.007       & 7.702 \textpm \ 0.069      & 7.517 \textpm \ 0.046      & 7.204 \textpm \ 0.082      \\
                          &                        & sMAPE (\%)                   & 11.694 \textpm \ 0.000 & 11.424 \textpm \ 0.088 & 10.626 \textpm \ 0.007 & 10.580 \textpm \ 0.029 \\ \cline{2-7} 
                          & \multirow{3}{*}{60min} & MAE                     & 6.408 \textpm \ 0.006      & 6.016 \textpm \ 0.040      & 5.672 \textpm \ 0.002      & 5.557 \textpm \ 0.011      \\
                          &                        & RMSE                    & 11.011 \textpm \ 0.007     & 10.128 \textpm \ 0.083     & 9.628 \textpm \ 0.003      & 9.438 \textpm \ 0.069      \\
                          &                        & sMAPE (\%)                   & 14.709 \textpm \ 0.027 & 13.624 \textpm \ 0.212 & 12.543 \textpm \ 0.048 & 12.335 \textpm \ 0.055 \\ \hline
\end{tabular}
\end{table*}
\label{subsect:network_traffic_results}

In order to evaluate the effectiveness of the integration approach, we examine the performance of the inception, \ac{seacnn}, \ac{mtgnn}, and \ac{seagnn} models on the network traffic datasets. As in the road traffic datasets, we perform the experiment 5 times with varying random seed numbers and compute the mean and standard deviation of the results for each evaluation metric that are summarized in \Tableref{tab:network_results}. As can be seen from the table, SEACNN provides slightly better results than the inception model in terms of \ac{mae} and \ac{rmse} metrics, especially for the MI-SMS dataset. The gain becomes more significant for larger prediction horizons, e.g., 60 minutes. Notably, \ac{seagnn} outperforms \ac{mtgnn} across all prediction horizons for two datasets, which demonstrates the effectiveness of the integration approach. The reason the accuracy gain on the network traffic datasets is not as significant as those on the road traffic datasets could be due to the limited size of the network datasets, which is of 61 days with a coarser granularity of 10 minutes as indicated in \Tableref{tab:exp_dataset}.

\section{Conclusions}
\label{sect:conclusions}
In this work, we describe a complete approach to integrate the time component into forecasting models in order to improve their performance. The approach consists of the auxiliary time feature extraction, the \ac{cnf}, and the \ac{lgf} modules. The first two modules are used to extract the seasonality patterns that are considered to be the global information of the time series. The \ac{lgf} module is then used to fuse the global information with the local information extracted by autoregressive neural networks. To demonstrate the effectiveness of the proposed approach, we implement two novel models, namely \ac{seacnn} and \ac{seagnn} which are the integration of the time component using the proposed approach into the inception baseline model and the state-of-the-art \ac{mtgnn} graph-based model. Experimental results on 2 road traffic datasets and 2 network traffic datasets indicate the effectiveness of the two novel models over their underline counterparts, demonstrating that the proposed approach is not only effective in convolution-based forecasting models but also in graph-based ones. Therefore, the integration approach is general and could be straightforwardly applied to any other forecasting models such as TraverseNet \cite{traversenet_2022} and Transformer-based Informer \cite{informer_2021}, which are also our short-term perspectives.

\bibliographystyle{IEEEtran}
%\bibliography{refs}

\appendix

\section*{Appendix A: Expressivity Evaluation via a Reconstruction Task}
\label{sect:appendix_a}

\begin{table}[h]
\centering
\caption{The loss of the reconstruction task}
\label{tab:reconstruction}
\begin{tabular}{@{}cccc@{}}
\toprule
Node & RFF & SIREN & Vanilla MLP \\ \midrule
20 & \textbf{0.3761 \textpm \ 0.0004} & 0.3850 \textpm \ 0.0021 & 0.4165 \textpm \ 0.0023 \\
35 & \textbf{0.3637 \textpm \ 0.0002} & 0.3710 \textpm \ 0.0008 & 0.3983 \textpm \ 0.0021 \\
56 & \textbf{0.4516 \textpm \ 0.0011} & 0.4640 \textpm \ 0.0008 & 0.5138 \textpm \ 0.0040 \\ \bottomrule
\end{tabular}%
\end{table}

\begin{figure}
    \centering
    \begin{minipage}[t]{\columnwidth}
    %\centerline{\includesvg[inkscapelatex=false,width=\textwidth]{figures/reconstruction_20.svg}}
    \centerline{\includegraphics[width=\textwidth]{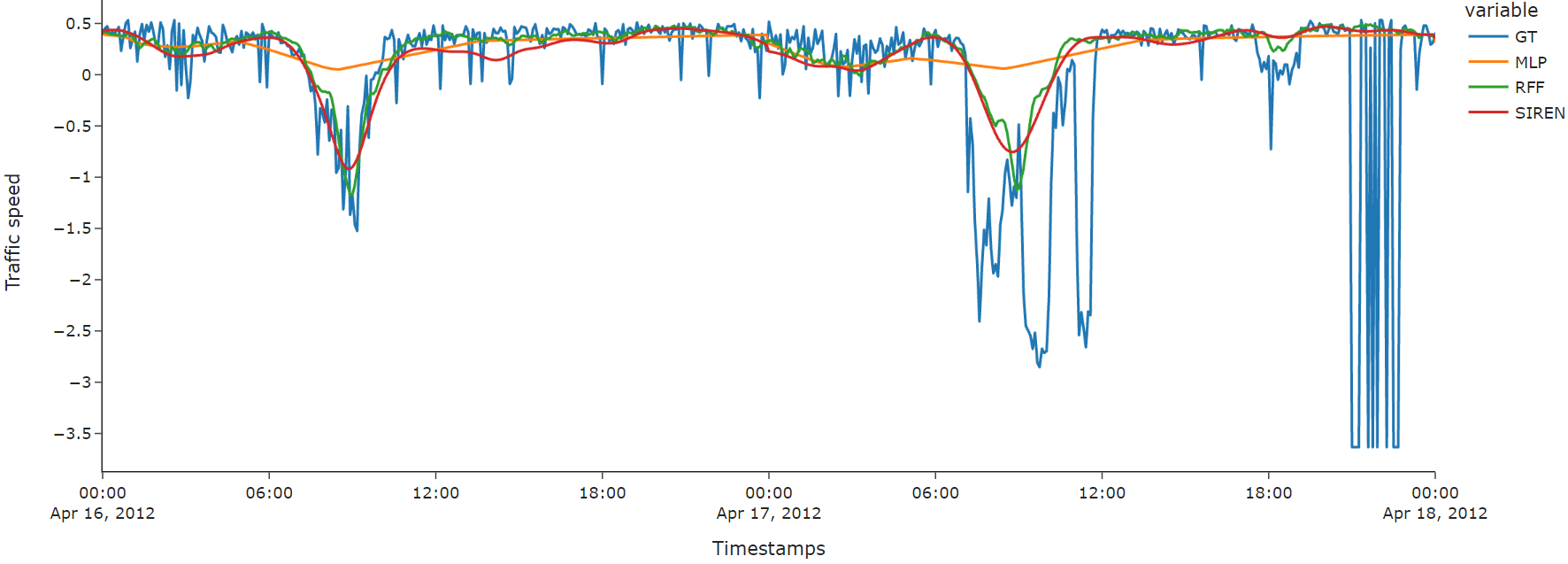}}
    %\centerline{(a) Node 20}
    \end{minipage}
    %\vfill
    
    \begin{minipage}[t]{\columnwidth}
    %\centerline{\includesvg[inkscapelatex=false,width=\textwidth]{figures/reconstruct_35.svg}}
    \centerline{\includegraphics[width=\textwidth]{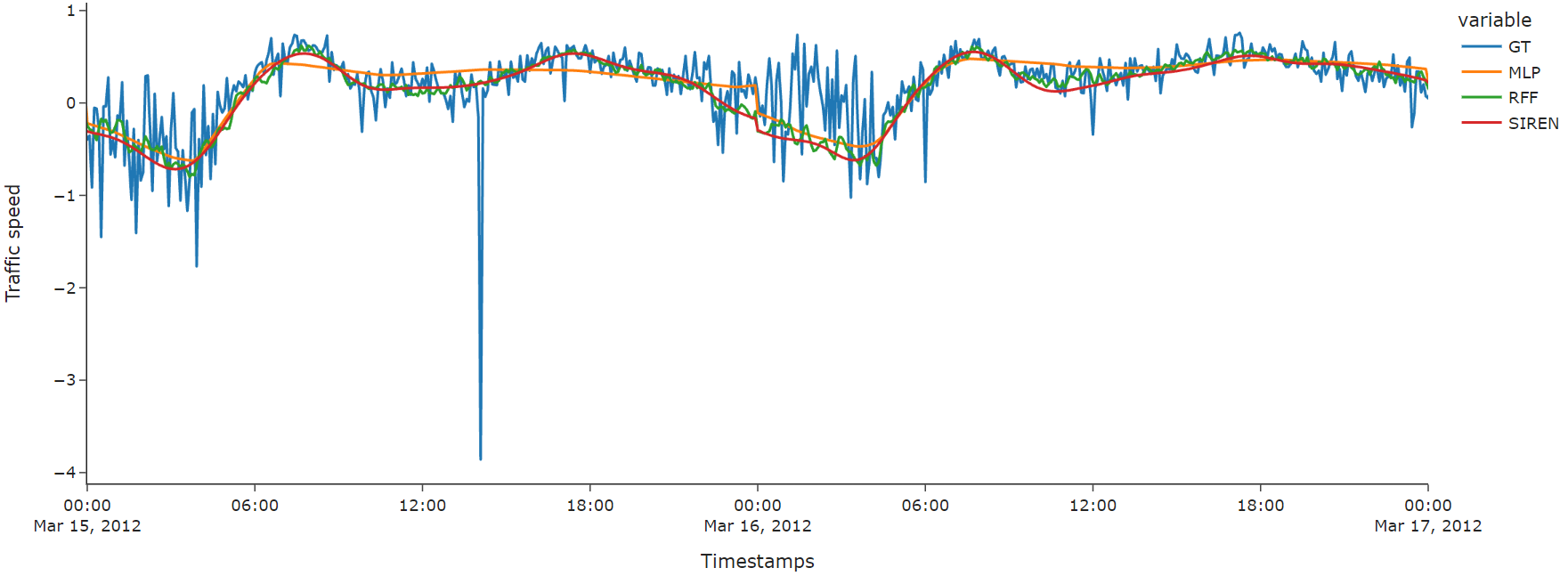}}
    %\centerline{(b) Node 35}
    \end{minipage}
    
    \begin{minipage}[t]{\columnwidth}
    %\centerline{\includesvg[inkscapelatex=false,width=\textwidth]{figures/reconstruct.svg}}
    \centerline{\includegraphics[width=\textwidth]{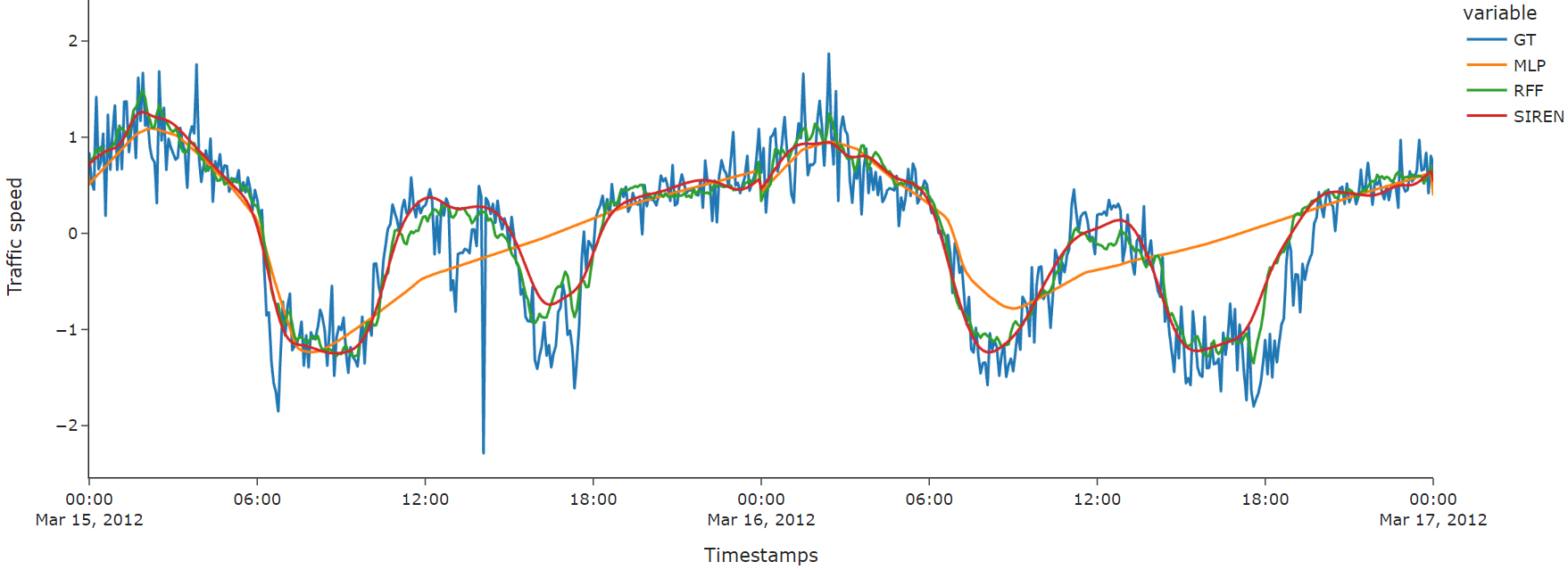}}
    %\centerline{(c) Node 56}
    \end{minipage}
    \caption{Illustration of the reconstruction of the time series for three nodes.}
    \label{fig:reconstruction}
\end{figure}

In order to evaluate the expressivity of the \ac{mlp}, we implement a coordinate-based \ac{mlp} of 1 hidden layer containing 128 hidden units to reconstruct a single univariate time series taken from the training set of the METR-LA dataset. Three univariate time series corresponding to 3 nodes of 20, 35 and 56 from 207 nodes are randomly selected for the experiment. We train the \ac{mlp} to minimize the \ac{mae} reconstruction loss. We evaluate with 3 variants of the \ac{mlp} including: RFF, SIREN \cite{vsitzmann} and the vanilla \ac{mlp}. SIREN is a \ac{mlp} network with a sine periodic activation function, which relates to the Time2Vec framework \cite{time2vec} in which the authors introduce a learnable embedding of the timestamps via trend and seasonality decomposition. 

\Tableref{tab:reconstruction} presents the reconstruction loss which is the average and standard deviation of \ac{mae}s of 3 runnings with different random seeds. The table shows that by enhancing the \ac{mlp} with the RFF, the best reconstruction results are achieved for all 3 nodes, followed by SIREN and then the vanilla \ac{mlp}. \Figref{fig:reconstruction} shows the reconstructed results compare to the ground truths. As can be seen from the figure, while the SIREN and RFF variants succeed in fitting the time series, the vanilla MLP tends to over-smooth out the reconstructed signals, exhibiting the spectral bias problem. Furthermore, despite the qualitative success in reconstruction, SIREN produced smoother results than the RFF. In this implementation, we choose 10 as the standard deviation of the Gaussian distribution in the RFF after some fine-tuning. In conclusion, using the reconstruction task of univariate time series, we identify the RFF as the best-parameterized network for the neural field. Therefore, we use the neural field implemented with RFF to extract the global features in the forecasting pipeline.

% Handling missing values

\end{document}